\documentclass[10pt, a4paper]{article}
\usepackage[table]{xcolor} % 引入宏包并使用table选项
\usepackage[final]{lrec-coling2024} % this is the new style

\usepackage{booktabs} 
\usepackage{natbib}
\usepackage{multibib}

\usepackage{IEEEtrantools}
\usepackage{graphicx}
\usepackage{amsmath}
\usepackage{amsthm}
\usepackage{booktabs}
\usepackage{algorithm}
\usepackage{algorithmic}
\usepackage{natbib} 
\usepackage{amssymb}
\usepackage{multirow}
\usepackage{boldline}
\usepackage{hhline}
\usepackage{amsmath}
\usepackage{pifont}
\usepackage{subfigure}
\usepackage{makecell}
\usepackage{mathrsfs}
\usepackage{utfsym}
\usepackage[normalem]{ulem}
\useunder{\uline}{\ul}{}
\usepackage[inline]{enumitem}
\usepackage{enumitem}
\usepackage{IEEEtrantools}
\usepackage{stmaryrd}
\usepackage{mathabx}

\makeatletter
\def\@mb@citenamelist{cite,citep,citet,citealp,citealt,citepalias,citetalias}
\makeatother
\newcites{languageresource}{~}

\usepackage{graphicx}
\usepackage{tabularx}
\usepackage{soul}
\usepackage{amsmath}
\usepackage{xstring}
\usepackage{hyperref}
 \definecolor{darkblue}{rgb}{0, 0, 0.5}
  \hypersetup{colorlinks=true, citecolor=darkblue, linkcolor=darkblue, urlcolor=darkblue}
\usepackage{array} %
\usepackage{color} %
\usepackage{soul} %
\usepackage{colortbl}
\usepackage{arydshln}%
\title{A Regularization-based Transfer Learning Method for Information Extraction via Instructed Graph Decoder}

\name{Kedi Chen, Jie Zhou, Qin Chen$^{*}$\thanks{$^*$ Corresponding author.}, Shunyu Liu, Liang He} 

\address{
         School of Computer Science and Technology, East China Normal University, Shanghai, China \\
         \{kdchen,71265901045\}@stu.ecnu.edu.cn \{jzhou, qchen, lhe\}@cs.ecnu.edu.cn\\}

\abstract{
Information extraction (IE) aims to extract complex structured information from the text. Numerous datasets have been constructed for various IE tasks, leading to time-consuming and labor-intensive data annotations. 
Nevertheless, most prevailing methods focus on training task-specific models, while the common knowledge among different IE tasks is not explicitly modeled. Moreover, the same phrase may have inconsistent labels in different tasks, which poses a big challenge for knowledge transfer using a unified model.
In this study, we propose a regularization-based transfer learning method for IE (\texttt{TIE}) via an instructed graph decoder. 
Specifically, we first construct an instruction pool for datasets from all well-known IE tasks, and then present an instructed graph decoder, which decodes various complex structures into a graph uniformly based on corresponding instructions. In this way, the common knowledge shared with existing datasets can be learned and transferred to a new dataset with new labels. Furthermore, to alleviate the label inconsistency problem among various IE tasks, we introduce a task-specific regularization strategy, which does not update the gradients of two tasks with `opposite direction'.
We conduct extensive experiments on 12 datasets spanning four IE tasks, and the results demonstrate the great advantages of our proposed method. % that our \texttt{TIE} method achieves new state-of-the-art performance.
\\ \newline  }

\begin{document}

\maketitleabstract

\section{Introduction}
Information extraction (IE) is a task to extract structured information (e.g., entities, relationships, and events) from textual data. IE encompasses many subtasks, including named entity recognition (NER), relation extraction (RE), event extraction (EE), and aspect-based sentiment analysis (ABSA). 
It is a challenging task due to the large label space and complex structure of various tasks.

Existing researches in IE can be categorized into two main classes: task-specific and unified models. 
Task-specific models \citep{chen-etal-2023-learning,wadhwa-etal-2023-revisiting,you-etal-2023-jseegraph,ma-etal-2023-amr} entail designing a unique structure for each individual task. 
These independent architectures require higher development costs and resources.  
% Despite achieving high accuracy, task-specific methods impede knowledge sharing among tasks.
Unified models, on the other hand, deploy a cohesive framework to address multiple tasks simultaneously. 
% Therefore, the adoption of a unified model is imperative. 
Presently, unified models predominantly employ a generative framework, translating extraction tasks into a sequence generation architecture  \citep{DBLP:conf/acl/0001LDXLHSW22,DBLP:conf/acl/0009C23}. 
Although the frameworks are structurally unified, most of the previous methods \citep{DBLP:conf/acl/0001LDXLHSW22,DBLP:conf/acl/0001SLZHQ23} merely finetune the models on the target dataset, disregarding the common knowledge within numerous existing IE datasets, 
% Numerous datasets are available for each subtask, 
including ACE 2005 \citep{walker2006ace}, CoNLL03 \citep{DBLP:conf/conll/SangM03}, 16-res \citep{DBLP:conf/semeval/PontikiGPAMAAZQ16}, and others. 
% complex task-invariant and task-specific knowledge between the existing IE datasets.
% In order to fully leverage the extensive overlapping knowledge in IE, transfer learning is a viable solution.
This paper emphasizes the acquisition of shared knowledge across these tasks and datasets.

\setulcolor{cyan}
\definecolor{lightgray}{gray}{0.9}
\definecolor{deepgray}{gray}{0.5}
\begin{table}[t] 
\centering 
\scriptsize{
    \begin{tabular}{l|p{1.3cm}<{\centering} p{2.1cm}<{\centering}}
        % \hline
        \hline
        \textbf{CoNLL03 example} & \textbf{label} & \textbf{mention}  \\
        \hline
       \multirow{4}{*}{\parbox[t]{3cm}{VICORP \textcolor{blue}{restaurants$^{\left[\text{ORG}\right]}$} names Sabourin CFO.}}
        & PER & -  \\
        & \cellcolor{deepgray}ORG &  \cellcolor{lightgray}restaurants \\
        & LOC & -    \\
        & ... & ...    \\
        \hline
        \hline
        \textbf{ACE05-Rel example} & \textbf{label} & \textbf{mention}  \\
        \hline
        \multirow{6}{*}{\parbox[t]{3cm}{The explosion comes after a bomb exploded at a \textcolor{blue}{\ul{restaurant}$^{\left[\text{FAC}\right]}$} in \textcolor{blue}{\ul{Istanbul}$^{\left[\text{GPE}\right]}$}, leading to damage but no injuries.}}
                                & PER & -   \\
                                & ORG & -  \\
                                & \cellcolor{deepgray}\text{FAC}  & \cellcolor{lightgray}restaurant  \\
                                &...  & ...  \\
                                & PartWhole$^{\dagger}$  & restaurant,Istanbul  \\
                                &...  & ... \\
        \hline
        \hline
        \textbf{ACE05-Evt example} & \textbf{label} & \textbf{mention}  \\
        \hline
        \multirow{5}{*}{\parbox[t]{3cm}{He was a segregationist who once closed a \textcolor{red}{\ul{restaurant}%$^{\left[ORG\right]}$
        } he \textcolor{purple}{\ul{owned}$^{\left[\text{Trig}\right]}$} rather than served African-Americans.}}
                                & Trig & owned   \\
                                & \cellcolor{deepgray}ORG$^{\dagger}$ & \cellcolor{lightgray}  owned,restaurant \\
                                & Place$^{\dagger}$  & -  \\
                                &PER$^{\dagger}$  & -  \\
                                & ...  & ... \\
        \hline
        \hline
        \textbf{16-res example} & \textbf{label} & \textbf{mention} \\
        \hline
        \multirow{5}{*}{\parbox[t]{3cm}{The Petrus and Vonglas’s tiny \textcolor{blue}{\ul{restaurant}$^{\left[\text{Asp}\right]}$} is as \textcolor{blue}{\ul{cozy}$^{\left[\text{Exp}\right]}$} as it gets, with that certain Parisian flair.}}
                                & Expression & cozy  \\
                                & \cellcolor{deepgray} Aspect & \cellcolor{lightgray} restaurant \\
                                & Positive$^{\dagger}$  & cozy,restaurant  \\
                                & Negative$^{\dagger}$  & -  \\
                                & Neutral$^{\dagger}$  & -  \\
        \hline

    \end{tabular}
    }
\caption{An example of inconsistent annotations among different subtasks. $^{\dagger}$ means the relation labels. A relation label exists between two words with underscores.}
\label{tab:res_real_data_wide}
\vspace{-3mm}
\end{table}

Nevertheless, several challenges persist in knowledge transfer across distinct IE tasks.
\textbf{First}, the datasets originate from various IE tasks, resulting in substantial diversity in data structures. More specifically, 1) Two datasets from the same subtask may exhibit distinct entity or relation types. Thus, the target datasets may contain new label classes that do not occur in the source datasets; 2) Although two entity or relation types are semantically similar in different datasets, they are labeled with different names. For instance, the relation types `OrgBased\_In' in CoNLL04 and `PART-WHOLE' in ACE05-Rel share the same semantic meaning, signifying `locate in'; 3) Furthermore, certain labels encompass the meanings of multiple other labels. `MISC' of CoNLL03 is applied to label diverse miscellaneous entities. ACE05-Rel utilizes `GEN-AFF' to denote generic affiliations without specific references. These labels with vague semantics significantly influence the model's learning process.
\textbf{Second}, in different datasets pertaining to distinct IE subtasks, the same phrase may have inconsistent labels owing to various annotation guidelines. As depicted in Table~\ref{tab:res_real_data_wide}, the phrase `restaurant' in several datasets related to different IE subtasks exhibits different annotation information, including `ORG', `FAC', `Aspect', etc. 
This discrepancy introduces conflicts in the comprehension of the phrase for various IE subtasks.

% To address the above problems, we propose a gradient-regularization instruction learning model, which can harness all information extraction datasets, utilizing datasets with diverse structures. First, we manually write instructions for each IE task and employ ChatGPT to perform paraphrasing, ensuring semantic diversity. We denotes all the instructions as IE Instruction Pool. Instructions can provide guiding text to help the model adapt to different sub tasks and mitigate the disparities, enhancing models' generality\citep{DBLP:conf/icml/JangKYKLLLS23}. Then, we input both the text and a randomly selected instruction from the corresponding dataset into the language model, reinforcing the representation of each word in the text with label words. During the process of transfer learning, we employ a continual learning regulazation-based method in the training phase on the source datasets. This approach only update the gradient `in the same direction', which aims to learn common features of IE sub tasks. ultimately for testing on the target dataset. We eventually finetune on the target dataset.

To address the aforementioned challenges, we introduce a regularization-based transfer learning method for IE (\texttt{TIE}) via an instructed graph decoder. 
First, we design an instructed graph decoder to learn task-shared knowledge by modeling the various formats of different IE tasks as a graph.
Then, we propose a task-specific regularization transfer strategy to resolve conflicting knowledge among tasks.
% An instruction-based graph decoder is designed to leverage a wide range of information extraction datasets, each characterized by diverse structures.
The instructed graph decoder consists of two parts:
1) Instruction pool, which contains manually crafted task-specific instructions for each dataset of different IE tasks. 
% We employ ChatGPT for paraphrasing to enhance semantic diversity in these instructions. 
% All the instructions are aggregated into a unified resource referred to as the `Instruction Pool'.
These instructions serve as guiding text to facilitate model's adaptation to different datasets and mitigate disparities, thereby enhancing the generalization capability; 
2) Graph decoder, which decodes various formats of different tasks into a unified graph structure with instructions.
% Subsequently, we input both the text and a randomly selected instruction from the corresponding dataset into the language model. This enhances the representation of each word in the text through associated label words.
The task-specific regularization strategy does not update the gradients of two tasks `in the opposite direction', for resolving conflicting knowledge across tasks, ultimately preparing the model for testing on the target dataset.
The experimental results demonstrate that our approach achieves state-of-the-art in most of the IE datasets, even with improvements in data-scarce scenarios.

The main contributions of this paper can be summarized as follows:
\begin{itemize}%[leftmargin=*, align=left]
\item{We propose a \texttt{TIE} method that explicitly models the common knowledge from various IE datasets with an instructed graph decoder.}
\item{A task-specific regularization strategy is designed to help reduce the inconsistent labels or conflicts across diverse IE tasks, by not updating the gradients `in the opposite direction' during transfer learning.}
\item{Experiments on 12 datasets of four IE subtasks show the advantages of our proposed method. Moreover, our method is superior to the baselines on low-resource and few-shot scenarios\footnote{Our codes and datasets can be found in https://github.com/141forever/TransferUIE}.}
\end{itemize}

\section{Related Work}
\paragraph{Information Extraction}
Information extraction (IE), deriving structured information from unstructured source data, is an essential task in natural language processing (NLP). 
Information extraction contains several subtasks, such as named entity recognition \citep{DBLP:journals/csi/MarreroUSMB13}, relation extraction \citep{DBLP:conf/ccks/CuiLWY17}, event extraction \citep{DBLP:conf/emnlp/WaddenWLH19}, aspect-based sentiment analysis \citep{DBLP:journals/eswa/DoPMA19}, etc. For a period of time, researchers tend to work on these subtasks separately. 

In recent years, \citet{DBLP:conf/acl/0001LDXLHSW22} proposes a generative unified information extraction (UIE) model with structured extraction language and structural schema instructor. The generative paradigm generates too much redundant information and has poor completeness. The same authors then introduce a new framework USM \citep{DBLP:conf/aaai/Lou0DJLH0023} with token linking operations. However, USM brings unnecessary loss of time in both training and inference periods. The Plusformer architecture harnessed by \citet{DBLP:conf/acl/0001SLZHQ23} requires high algorithmic complexity, hence simplification is indispensable. \citet{DBLP:conf/acl/PingLGWZZZ23} converts IE tasks into span classification via the triaffine mechanism, but the reliability on syncretic complex-label datasets has not been validated.

With the advent of large language models (LLMs) \citep{DBLP:conf/acl/0009C23}, there have been significant changes in IE. 
% As new technologies like instruction tuning \citep{DBLP:conf/icml/JangKYKLLLS23}, chain of thought \citep{DBLP:conf/nips/Wei0SBIXCLZ22}, in-context learning \citep{DBLP:conf/emnlp/MinLHALHZ22}, reinforcement learning with human feedback \citep{DBLP:conf/nips/Ouyang0JAWMZASR22}, etc. gradually developing, the demand for computational resources is also increasing. 
ChatIE \citep{DBLP:journals/corr/abs-2302-10205} makes an initial attempt to use ChatGPT3.5 for performing information extraction tasks, through multi-turn conversations. The accuracy is not as precise as expected.  \citet{DBLP:journals/corr/abs-2304-11633,DBLP:journals/corr/abs-2305-14450} assess the information extraction capabilities of ChatGPT3.5 systematically, and find a gap between ChatGPT3.5 and SOTA results. InstructUIE presented by \citet{DBLP:journals/corr/abs-2304-08085} tests on 32 diverse information extraction datasets, employing language model FlanT5-11B \citep{https://doi.org/10.48550/arxiv.2210.11416} in a generative pattern. This method consumes a significant amount of computational resources, making the reproducibility of results challenging. 

Given the aforementioned issues, our method leverages a simple architecture and is capable of addressing complex annotations, finally achieving commendable performances in both small and large language model settings.

\paragraph{Gradient Regularization} is a regularization technique for deep learning, in order to improve generalization performance and prevent overfitting \citep{DBLP:journals/pr/LiS23}. This technique is widely deployed for coordinating the training of multiple tasks and preventing interference between them \citep{DBLP:conf/iclr/SahaG021,DBLP:conf/iclr/LinYFZ22}. A previous study illustrates that \textit{If the angle between the gradients of the current task and the past task is acute, it is less likely to increase the loss of the previous task} \citep{DBLP:conf/nips/Lopez-PazR17}. This finding serves as a critical theoretical foundation for our method, presenting a possibility that models can resolve inconsistent knowledge of different tasks.

\begin{figure*}[!t] 
\centering
    \includegraphics[width=1\textwidth]{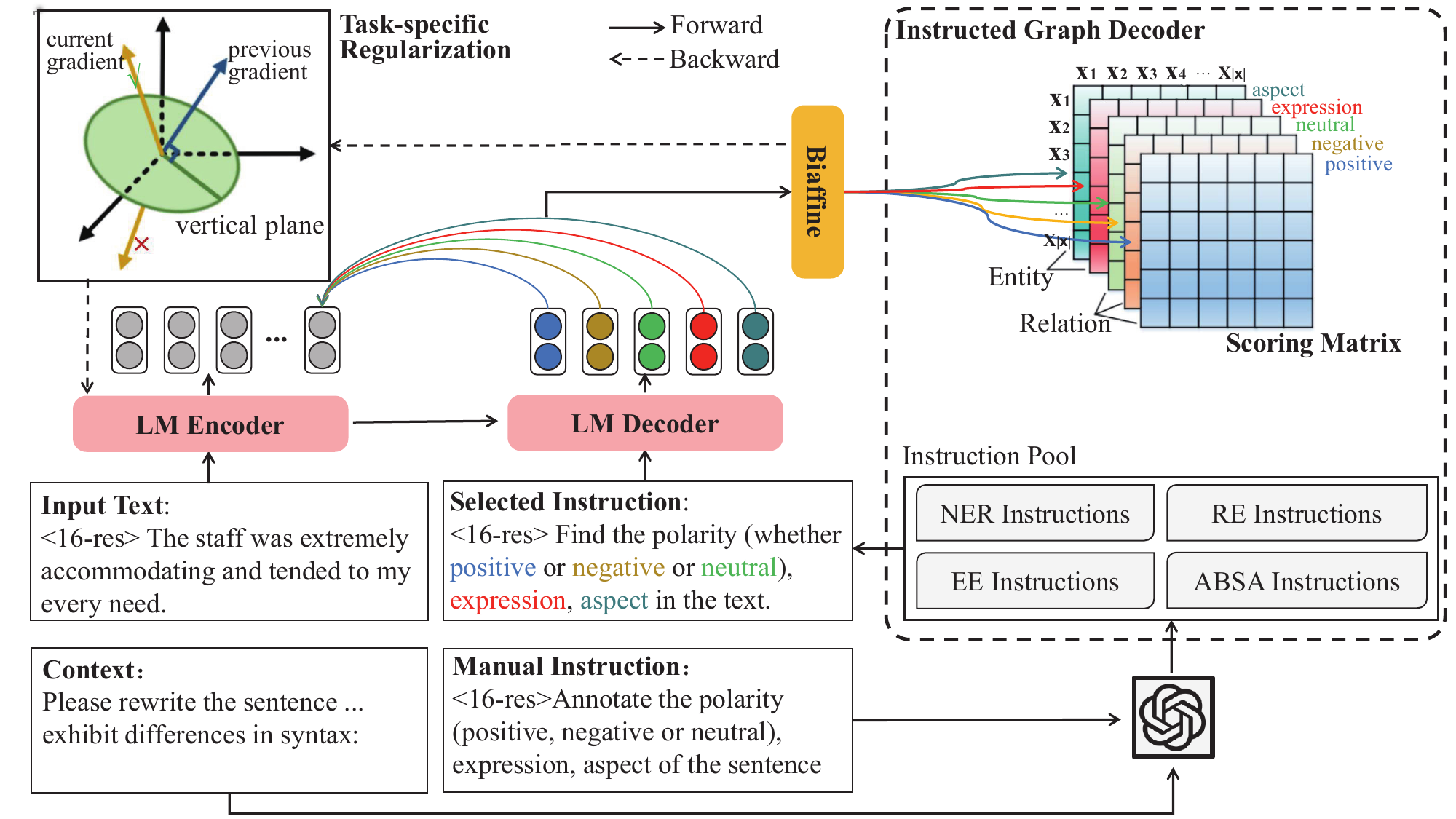} 
    \vspace{-5mm}
    \caption{The framework of our \texttt{TIE} method.} 
    \label{fig:main}
    \vspace{-2mm}
\end{figure*}

\paragraph{Transfer Learning for IE}
Transfer learning is an important approach to enhance the generalization of deep learning. The purpose of transfer learning is to enhance the performance of models within target domains by leveraging the knowledge from correlated source domains  \citep{DBLP:journals/pieee/ZhuangQDXZZXH21}. In the field of IE, many works indicate the superiority of transfer learning. When it comes to NER, \citet{DBLP:series/sci/BhatiaAC20} proposes a dynamic transfer network to learn sharing parameters between tasks. \citet{DBLP:conf/kdd/DiSC19} addresses the label sparsity problem of relation extraction in a real-world scenario. By combining variational information bottleneck into a model called SharedVIB which can search for structured common knowledge, \citet{DBLP:conf/coling/ZhouZCHH22} boosts the correlation between three event argument extraction tasks. However, these works merely focus on the knowledge among one IE subtask (intra). In contrast, our approach focuses on the inter-task transfer. Moreover, we explore to resolve the inconsistent labels or conflicts during transfer learning.

\section{Our Method}
In this section, we introduce the framework of our method (Figure \ref{fig:main}), which consists of two parts: instructed graph decoder and task-specific regularization.
% First, to model the various formats of different IE tasks, we design an instructed graph decoder to learn the task-shared knowledge among various datasets. 
% Then, we propose a task-specific regularization transfer strategy to reduce conflicting knowledge among IE subtasks.
First, we manually craft instructions for each dataset and utilize ChatGPT3.5 to paraphrase, forming an instruction pool.
Then, we present an instructed graph decoder to obtain the instruction-activating representations of the input text. It also learns common knowledge by modeling all the structured information with a graph represented by a token matrix. 
% we propose an instructed graph decoder based on an encoder-decoder structure. In particular, we input the text along with a random selected instructions, in order to achieve the label-sensitive graph structure.
Moreover, in order to alleviate the conflicts between various IE tasks, we present a task-specific regularization strategy that does not update the gradients `in opposite direction’ between source tasks during training on source datasets, and finally finetune on the target dataset.

\subsection{Task Definition}
We regard any single IE task as an instruction-activating span annotation mission on dataset $\mathcal{D}$. 
Given an instance $\left(\boldsymbol{x},\mathcal{E},\mathcal{R},\mathcal{I}\right)\in\mathcal{D}$, where $\boldsymbol{x}=\left(x_1,\dots,x_{|\boldsymbol{x}|}\right)$ is the input sentence with $|\boldsymbol{x}|$ tokens.
$\mathcal{E}$, $\mathcal{R}$, $\mathcal{I}$ denotes the set of entity types, the set of relation types and the set of instructions separately. 
Regarding the named entity recognition task, $\mathcal{R}=\emptyset$. 
As for the event extraction task, $\mathcal{E}=\mathcal{T}$ and $\mathcal{R}=\mathcal{A}$, where $\mathcal{T}$ regards the set of trigger types or event types, and $\mathcal{A}$ represents the set of argument roles. 
In reference to aspect-based sentiment analysis task, 
% which can also be known as sentimental tuple extraction task, 
$\mathcal{E}=\left\{\mathrm{Aspect}, \mathrm{Expression}\right\}$ and $\mathcal{R}=\left\{\mathrm{Positive}, \mathrm{Negative}, \mathrm{Neutral}\right\}$. 
We aim to achieve a scoring matrix $\mathbf{M}^{|x|*|x|*\left(|\mathcal{E}|+|\mathcal{R}|\right)}$, which can indicate the label of each span of the input sentence. 
$\mathbf{M}\left[i,j,k\right]=1$ means the span $(i,j)$ of the input sentence has label $k$.

\subsection{Instructed Graph Decoder}
In this module, we first use an instruction pool to translate the IE labels into instructions so that the model can learn the representations of class effectiveness and capture new label classes. 
Then, we apply a graph decoder based on instructions obtained from the instruction pool to decode the complex and various structures into a graph uniformly. 
We combine the instruction pool and the graph decoder together, referring to them as an instructed graph decoder.

\begin{table}[t] 
\centering 
\scriptsize{
        \small
    \begin{tabular}{c|p{5.3cm}<{\centering}}
       \hlineB{4}
        \textbf{Dataset} & \textbf{Instruction Example} \\
         \hline
       \multirow{3}{*}{ACE04} & Identify entities (organization, person, vehicle, geographic, location, weapon, facility) mentioned in the sentence.\\
       \hline
       \multirow{4}{*}{CoNLL04} & Explore the relationships work for, locate in, base in, live in, and kill someone between the entities location, organization, people and other. \\
       \hline
       \multirow{3}{*}{ACE05-Evt}& Locate the mentioned event types: acquit,..., trial hearing. Identify the argument types: adjudicator,..., victim. \\
       \hline
       \multirow{3}{*}{16-res} & Find the sentiment (positive, negative or neutral) of the sentence and identify the expression, aspect element. \\

        \hlineB{4}
    \end{tabular}
    }
\vspace{-2mm}
\caption{One instruction example for four datasets of different IE subtasks respectively.}
\label{tab:instruction example}
\vspace{-3mm}
\end{table}

\paragraph{Instruction Pool}
For various datasets or tasks, the label spaces are different. 
%Instead of regarding IE as a classifier,
We create a set of instructions for each dataset $\mathcal{D}$, all the instructions are referred to as an instruction pool. 
Each instruction contains all entity types $e \in \mathcal{E}$ and relation types $r \in \mathcal{R}$ of the corresponding dataset. 
In this way, the model can learn representations of similar labels and new classes. 
% All instructions from different datasets are referred to as an instruction pool. 

For each dataset, we first write an instruction manually. 
Take dataset 16-res of ABSA task as an example, given entity types $\mathcal{E}$ and relation types $\mathcal{R}$, the instruction we designed artificially is: 
``\textit{Annotate the polarity (\underline{positive}, \underline{negative} or \underline{neutral}), \underline{expression}, \underline{aspect} of the sentence.}"
Then, to improve the diversity of the instruction, we adopt ChatGPT3.5 to augment the instruction. Specifically, the complete prompt input into ChatGPT3.5 is: 
``\textit{Please rewrite the following sentence several times and make sure the rewritten sentences exhibit significant differences in syntax, compared to the original sentence: Annotate the polarity (\underline{positive}, \underline{negative} or \underline{neutral}), \underline{expression}, \underline{aspect} of the sentence.}" More details of the construction of the instruction pool are shown in Appendix~\ref{app: instruction pool}. 
% Finally, we obtain the diverse instructions via ChatGPT:
% \begin{itemize}[leftmargin=*, align=left]
%     \item \textit{Get the polarity (whether \underline{positive} or \underline{negative} or \underline{neutral}) of the sentence, and find the \underline{expression} text, \underline{aspect} text.}
%     \item  \textit{Retrieve the sentiment (\underline{positive},\underline{negative}, or \underline{neutral}) of the sentence and identify the corresponding content as either \underline{expression} text, \underline{aspect} text.}
% \end{itemize}\

Instructions of the other datasets can be obtained in the same way. 
We provide one instruction example for four datasets of different IE subtasks respectively in Table~\ref{tab:instruction example}. For more examples, please refer to Appendix~\ref{app: examples of the instrutions}.

%Please refer to Appendix~\ref{appd: instructions} for the instructions corresponding to each dataset.

\paragraph{Graph Decoder}
To capture the complex structures of various IE tasks, we design a graph decoder to decode all the structured information as a graph.
Given an input text with $|x|$ tokens $\boldsymbol{x}=\left[x_1,\dots,x_{|x|}\right]$, we harness T5 \citep{2020t5,https://doi.org/10.48550/arxiv.2210.11416} series to model the sentences and instructions. It is an adaptable encoder-decoder pre-trained language model (PLM) $\mathcal{M}=\left[\mathcal{M}_{enc},\mathcal{M}_{dec}\right]$ designed to tackle many NLP tasks.

We first use the encoder of PLM to obtain the hidden representation of input sentence $\boldsymbol{x}$ as follows.
\begin{equation}
\mathbf{H}^{enc}_{x} = \left[\mathbf{h}_x^1,\dots,\mathbf{h}_x^{|x|}\right] = \mathcal{M}_{enc}(\left[x_1,\dots,x_{|x|}\right]) 
\end{equation}
where $\mathbf{H}^{enc}_{x} \in \mathbb{R}^{|x|*d}$, $d$ is the dimension of hidden layers.

Next, to model the interaction between the sentence and the instruction, the decoder part of the PLM is leveraged to get sentence-aware instruction representation. 

As mentioned earlier, we construct several diverse instructions for each dataset of different IE subtasks, which make up an instruction pool. For each sample, we randomly select an instruction corresponding to the dataset. The selected corresponding instruction is denoted as $\boldsymbol{u}$ with length $|u|$ from the instruction pool and is inputted into the decoder. 
% The decoder part of the PLM is employed to get the feature of each label slot:
\begin{equation}
\mathbf{H}^{dec}_{u} = \left[\mathbf{h}_u^1,\dots,\mathbf{h}_u^{|u|}\right] = \mathcal{M}_{dec}(\mathbf{H}^{enc}_{x};\boldsymbol{u}) 
\end{equation}
where $\mathbf{H}^{dec}_{u} \in \mathbb{R}^{|u|*d}$, $d$ is the size of the hidden dimension. 
% and $K=|\mathcal{E}|+\mathcal{R}|$ label slots 

We can then achieve the representations of $K=|\mathcal{E}|+|\mathcal{R}|$ label slots, $\mathbf{H}_{slot} = \left\{\mathbf{h}_u^{\mathrm{slot\_index}(i)} \right\}^K_{i=1}$, where $\mathrm{slot\_index}\left(i\right)$ is the index of the $i$-th label slot in the instruction. Each $\mathbf{h}_u^{\mathrm{slot\_index}(i)} \in \mathbf{H}^{dec}_{u}$ and $\mathbf{H}_{slot} \in\mathbb{R}^{K*d} $.

Finally, to obtain the label-sensitive text representation $\mathbf{H}_{x} = \left[\mathbf{h}^1,\dots,\mathbf{h}^{|x|}\right]$, we deploy attention operations \citep{DBLP:conf/nips/VaswaniSPUJGKP17} to $\mathbf{H}^{enc}_{x}$ and $\mathbf{H}_{slot}$.
\begin{align}
    % \mathbf{h}^j &= \sum^K_{i=1} \mathbf{h}_u^j * \frac{e^{\langle \mathbf{h}_u^j, \mathbf{h}_u^i \rangle}}{\sum e^{\langle \mathbf{h}_u^j, \mathbf{h}_u^i \rangle}}
    \mathbf{H}_{x} = \mathrm{Softmax}(\mathbf{H}^{enc}_{x}\mathbf{W}_1(\mathbf{H}_{slot}\mathbf{W}_2)^T)\mathbf{H}_{slot}\mathbf{W}_2
\end{align}
$\mathbf{W}_1$,$\mathbf{W}_2 \in \mathbb{R}^{d*d}$ are learnable parameters.
% where $\mathbf{h}^j \in \mathbf{H}_{x}$, $\mathbf{h}_u^j \in \mathbf{H}_{u}^{enc}$ and $\mathbf{h}_u^i \in \mathbf{H}_{slot}$. $\sum$ between vectors represents element-wise addition and $\langle  ,\rangle$ means dot product operations.

At last, we represent the graph structure of the tokens using a matrix and calculate the scoring matrix in a biaffine way \citep{DBLP:conf/acl/BarnesKOOV20,DBLP:conf/acl/0001SLZHQ23} with multilayer perceptron (MLP).
\begin{align}
    \mathbf{H}^{head}_x &= MLP_{head}(\mathbf{H}_x) \\
    \mathbf{H}^{tail}_x &= MLP_{tail}(\mathbf{H}_x) \\
    \mathbf{M}_x\left[i,j\right]      =   & (\mathbf{H}^{head}_x\left[i\right])^T \mathbf{W}_3 \mathbf{H}^{tail}_x\left[j\right] \notag \\
                        &+\mathbf{W}_4 [\mathbf{H}^{head}_x\left[i\right];\mathbf{H}_x^{tail}\left[j\right]] \\
    \mathbf{M}         &=MLP_{score}(\mathbf{M}_x)
\end{align}
where $\mathbf{H}^{head}_x, \mathbf{H}^{tail}_x \in \mathbb{R}^{|x|*d}$, $\mathbf{M}_x, \mathbf{M} \in \mathbb{R}^{|x|*|x|*K}$, $\mathbf{W}_3 \in \mathbb{R}^{d*K*d}$,$\mathbf{W}_4 \in \mathbb{R}^{K*2d}$. $[;]$ means the concatenation between two vectors.

\subsection{Task-Specific Regularization}
Training all IE datasets within a unified model can facilitate the acquisition of shared knowledge across diverse datasets. Nevertheless, differences in task definitions and annotation guidelines can result in inconsistent labels. These variations in task-specific knowledge significantly impact the effectiveness of transfer learning. Consequently, we introduce a task-specific regularization technique, aimed at mitigating the influence of task-specific knowledge. 
% In order to learn common features and common knowledge from source tasks to assist in the training of the target task, we employed a transfer learning approach. For example. if our target dataset is 16-res of ABSA task, we first train our model on the other 11 datasets, and then finetune on dataset 16-res. These two steps can be referred to as `Common Knowledge Learning' and "Target Transfer Learning" respectively.

\paragraph{Task-Specific Knowledge Unlearning} During this step, we design a task-specific regularization method to unlearn conflicting knowledge. 
% Various tasks have different data structures, but they still share common features and knowledge across them. 
Particularly, to resolve the conflicting knowledge
among tasks, we do not update the model when the gradients of two consecutive tasks are `in an opposite direction'. That is, parameters are updated only when the angle between the current gradient and the previous time step's gradient is less than 90 degrees, otherwise, no update is performed \citep{DBLP:conf/nips/Lopez-PazR17}. We ensure that all data within one batch came from the same dataset, while the data in the two adjacent batches come from two different datasets of different IE tasks. Under the circumstances, the neighboring gradients signify the updating directions for different tasks. The angle is determined by the sign of the dot product result between two consecutive gradients. 
\begin{equation}
\textbf{Update} = 
\begin{cases} 
\mathrm{True}, & \text{if } \langle  \mathbf{g}_t , \mathbf{g}_{t-1} \rangle > 0 \\
\mathrm{False}, & \text{otherwise} 
\end{cases}
\end{equation}
where $\mathbf{g}_t$ and $\mathbf{g}_{t-1}$ are the gradients of the adjacent two batches respectively. If $\textbf{Update}=\mathrm{Fasle}$, we freeze the parameters of the corresponding layers.

Then, we finetune the transferred model directly on the target dataset. We take advantage of binary cross-entropy (BCE) loss to optimize the model.
% \begin{align}
% \mathcal{L}(\mathbf{M}, \mathbf{G}) &= -\frac{1}{|x|*|x|*K} * \notag\\
% \sum_{i=1}^{L} \sum_{j=1}^{L} \sum_{r=1}^{k} &（\mathbf{G}\left[i,j,r\right]\log(\mathbf{M}\left[i,j,r\right]) +\\
%  (1 - \mathbf{M}\left[i,j,r\right])& \log(1 - \mathbf{G}\left[i,j,r\right]) \notag
% \end{align}
% \begin{equation}
% BCEloss\left(a,b\right)=-a·log\left(b\right) - \left(1 - a\right)·log\left(1 - b\right) 
% \end{equation}
\begin{equation}
\mathcal{L}\left(\mathbf{M}, \mathbf{G}\right) = \sum_{i=1}^{|x|} \sum_{j=1}^{|x|} \sum_{r=1}^{K}\mathrm{BCE}\left(\mathbf{G}\left[i,j,r\right],\mathbf{M}\left[i,j,r\right]\right) 
\end{equation}
where $\mathbf{G}$ is the ground truth matrix, $K$ denotes the number of label slots and $r$ is the index of each label.

The task-specific regularization strategy is applied for updating parameters of the whole model, including the instructed graph decoder, which aims to preserve common knowledge and resolve conflicting knowledge among tasks during pre-training. Then, we finetune the whole pre-trained model including the instructed graph decoder on the target dataset.
% \begin{figure*}[!ht]
% \begin{center}
% %\fbox{\parbox{6cm}{
% %This is a figure with a caption.}}
% % old picture \includegraphics[scale=0.5]{lrec2020W-image1.eps} 
% \includegraphics[scale=0.5]{torino2024-banner.jpg} 
% \caption{The framework of our method.}
% \label{fig:framerwork}
% \end{center}
% \end{figure*}

\section{Experimental Setups}
% In this section, we conduct various experiments to validate the effectiveness of our model.

\subsection{Datasets}
For the main experiment, we follow the previous works \citep{DBLP:conf/acl/0001LDXLHSW22} and select 12 IE benchmark datasets of 4 IE subtasks: NER, RE, EE and ABSA. The specific datasets include: ACE04 \citep{mitchell2005ace}, ACE05-Ent \citep{walker2006ace}, CoNLL03\citep{DBLP:conf/conll/SangM03}; CoNLL04 \citep{DBLP:conf/conll/RothY04}, ACE05-Rel \citep{walker2006ace}, SciERC \citep{DBLP:conf/emnlp/LuanHOH18}; ACE05-Evt \citep{walker2006ace}, CASIE \citep{DBLP:conf/aaai/SatyapanichFF20}; 14-res \citep{DBLP:conf/semeval/PontikiGPPAM14}, 14-lap \citep{DBLP:conf/semeval/PontikiGPPAM14}, 15-res \citep{DBLP:conf/semeval/PontikiGPMA15}, 16-res \citep{DBLP:conf/semeval/PontikiGPAMAAZQ16}. According to our transfer learning configuration, we pre-train the model to learn the common knowledge and alleviate the inconsistency on 11 source datasets, and then fintune on the target dataset. 
The specific information about these datasets can be found in Table \ref{tab:statistics}.

For data-scarce scenarios, in order to make a fair comparison with \citet{DBLP:conf/acl/0001LDXLHSW22}, we adopt CoNLL03, CoNLL04, ACE05-Evt and 16-res datasets in few-shot and low-resource settings. 

\subsection{Metrics}
We employ Micro-F1 to assess the model's performance across various IE tasks.

\begin{itemize}[leftmargin=*, align=left]
    \item \textbf{Entity (Ent.F1).} An entity is correct if its entity type and span offsets both match a ground truth. 
    \item \textbf{Relation (Rel.F1).} A relation is correct if its type, along with the types and span offsets of both head and tail entities all match a ground truth. 
    \item \textbf{Event trigger (Trig.F1).} An event trigger is correct if its offsets and the event type both match a ground truth.
    \item \textbf{Event argument (Arg.F1).} An event argument is correct if its offsets, role type and event type all match a ground truth.
    \item \textbf{Sentiment Triplet (Senti Trip.F1).} We actually conduct an aspect sentiment triplet extraction (ASTE) task, so a sentiment triplet is correct if its offsets of expression (opinion), offsets of aspect and the sentimental polarity all match a ground truth.
\end{itemize}

\begin{table}[t] 
\centering 
\scriptsize
\setlength{\tabcolsep}{1.0mm}{
        \small
    \begin{tabular}{l|*{3}{l}*{3}{c}}
       \hlineB{4}
        \textbf{Dataset} & \textbf{\#Train} & \textbf{\#Dev} & \textbf{\#Test} & \textbf{|Ent|} & \textbf{|Rel|} &\textbf{|Evt|} \\
         \hline
       ACE04 & 6,297 & 742 & 824 & 7 & - & - \\
       CoNLL03 & 14,041 & 3,250 & 3,453 & 4 & - & -\\
       ACE05-Ent & 7,178 & 960 & 1,051 & 7 & - & -  \\
         \hline
        ACE05-Rel & 10,051 & 2,424 & 2,050 & 7 & 6 & -\\
        CoNLL04 &  922 & 231 & 288 & 4 & 5 & -\\
        SciERC & 1,861 & 275 & 551 & 6 & 7 & - \\
       \hline
        ACE05-Evt & 19,204 & 901 & 676 & - & - & 33\\
        CASIE & 5,235 & 1,115& 2,121 & - & - & 5\\
         \hline
        14-res & 1,266 & 310& 492 & 2 & 3 & -\\
        14-lap & 906 & 219& 328 & 2 & 3 & -\\
        15-res & 605 & 148& 322 & 2 & 3 & -\\
        16-res & 857 & 210& 326 & 2 & 3 & -\\
        \hlineB{4}
    \end{tabular}
    }
\caption{Dataset statistics. $\#$ means the number of instances, and |*| is the number of categories of the corresponding dataset.}
\label{tab:statistics}
\end{table}

\begin{table*}[t!]  
\renewcommand\arraystretch{1.4} 
\centering 
 \scriptsize
% \resizebox{\textwidth}{!}{
\setlength{\tabcolsep}{0.3mm}{
\begin{tabular}{l|*{3}{c}|*{3}{c}|*{4}{c}|*{4}{c}}
 \hlineB{4}
& \textbf{ACE04} & \textbf{ACE05-Ent} & \textbf{CoNLL03}
& \textbf{ACE05-Rel} & \textbf{CoNLL04} & \textbf{SciERC} 
& \multicolumn{2}{c}{\textbf{ACE05-Evt}} & \multicolumn{2}{c|}{\textbf{CASIE}}  
& \textbf{14-res} & \textbf{14-lap} & \textbf{15-res} &  \textbf{16-res} \\
& \multicolumn{3}{c|}{Ent.F1}
& \multicolumn{3}{c|}{Rel.F1} 
&  Trig.F1 & Arg.F1 &  Trig.F1 & Arg.F1
& \multicolumn{4}{c}{Senti Trip.F1} \\
\hline
BERT-base    & 84.09 & 84.63 & - & - & - & - & - & - & - & - & 71.85 & 59.38 & 63.27 & 70.26 \\
UnifiedNER  & 84.22 & 82.31 & 92.88 & - & - & - & - & - & - & - & - & - & - & - \\
NERGraph       & 86.31 & 85.11 & - & - & - & - & - & - & - & - & - & - & - & - \\
PURE       & - & - & - & 63.90 & - & \underline{35.60} & - & - & - & - & - & - & - & - \\
DEGREE    & - & - & - & - & - & - & 70.90 & 56.30 & - & - & - & - & - & - \\
BDTF    & - & - & - & - & - & - & - & - & - & - & 74.35 & 61.74 & 66.12 & 72.27 \\
\hdashline
TANL    & - & 84.90 & 91.70 & 63.70 & 71.40 & - & 68.40 & 47.60 & - & - & - & - & - & - \\
UIE    & 85.69 & 83.88 & 91.94 & 62.73 & 73.48 & 35.35 & 71.33 & 50.62 & 69.14 & 58.56 & 72.55 & \textbf{62.94} & 64.41 & 72.86 \\
ChatGPT3.5  & - & - & 67.20 & 40.50 & - & 25.90 & 15.50 & 30.90 & - & - & 41.50 & 33.17 & 38.89 & 47.67 \\ 
 InstructUIE {\tiny ({FlanT5-11B})}  & - & \underline{86.66} & \underline{92.94} & - & - & - & \textbf{77.13} & \textbf{72.94} & 67.80 & \underline{63.53} & - & - & - & - \\ 
\hline
% \midrule
\texttt{TIE} {\tiny ({T5-base})} & \underline{87.59} & 86.42 & 92.92 & \underline{64.44} & \underline{73.58} & 34.41 & 73.09 & 56.71 & \underline{74.43} & 63.14 & \underline{75.69} & 60.36 & \underline{66.78} & \textbf{75.17} \\
\texttt{TIE} {\tiny ({FlanT5-3B})} & \textbf{88.86} & \textbf{87.74} & \textbf{93.17} & \textbf{64.45} & \textbf{74.32} & \textbf{40.90} & \underline{74.89} & \underline{63.30} & \textbf{75.38} & \textbf{66.99} & \textbf{76.97}& \underline{62.04} & \textbf{66.84} & \underline{74.05} \\
% \hline
% - Transfer     & 87.44 & 85.11 & 92.15 & 63.69 & 73.49 & 34.11 & 72.85 & 55.11 & 73.57 & 62.38 & 73.95 & 59.69 & 66.61 & 73.79 \\
% - Instruction & 86.18 & 84.22 & 91.67 & 62.87 & 71.71 & 32.79 & 71.97 & 52.19 & 68.78 & 57.77 & 70.08 & 58.55 & 64.32 & 72.93 \\
% - Regulation & 86.14 & 85.95 & 91.86 & 64.09 & 73.46 & 33.87 & 72.64 & 55.32 & 74.12 & 62.53 & 69.78 & 56.68 & 61.73 & 73.06 \\
 \hlineB{4}
    \end{tabular}}
\vspace{-2mm}
\caption{Main results of \texttt{TIE} and the baselines. The upper part and the middle part are task-specific and unified methods respectively. The best result of each dataset is bolded, and the second-best is underlined.}
\label{tab:main}
\vspace{-3mm}
\end{table*}

\subsection{Baselines}
To validate the effectiveness of our method, we select several task-specific models and four unified models as baselines, compared with our approach. 

These task-specific methods are shown as follows.
\begin{itemize}%[leftmargin=*, align=left]
\item{\textbf{BERT-base}\footnote{https://huggingface.co/google-bert/bert-base-uncased} \citep{DBLP:conf/naacl/DevlinCLT19} is the most famous PLM for many nlp tasks. The results of ACE04 and ACE05-Ent are copied from \citet{DBLP:conf/acl/PengLZ0Z23}, which replaces the backbone of UIE \citep{DBLP:conf/acl/0001LDXLHSW22} with BERT-base. The results of four ABSA tasks are from \citet{DBLP:conf/acl/XuCB20}. It is a span-level method, considering the interaction between the spans of targets and opinions.}
\item{\textbf{UnifiedNER} \citep{DBLP:conf/acl/YanGDGZQ20} utilizes a seq2seq framework for three NER datasets. Given the similarity, we select the Span setting.}
\item{\textbf{NERGraph} \citep{DBLP:conf/acl/WanR0022} treats a sentence as a graph, applying graph convolutional network (GCN) for encoding.}
\item{\textbf{PURE} \citep{DBLP:conf/naacl/ZhongC21} works on two independent encoders and solely uses the entity model to construct the relation model.}
\item{\textbf{DEGREE} \citep{DBLP:conf/naacl/HsuHBMNCP22} manually designs prompts to help event extraction task.}
\item{\textbf{BDTF} \citep{DBLP:conf/emnlp/ZhangYLLCD0X22} is a boundary-driven table filling (BDTF) approach for ABSA tasks.}
\end{itemize}

Here, we introduce the four unified models.
\begin{itemize}%[leftmargin=*, align=left]
\item{\textbf{TANL} \citep{DBLP:conf/iclr/PaoliniAKMAASXS21} is an early-stage unified information extraction model.}
\item{\textbf{UIE} \citep{DBLP:conf/acl/0001LDXLHSW22} is a popular unified information extraction framework in the generative way. To ensure consistency in the backbone, we chose results from the official UIE-base model.}
\item{\textbf{ChatGPT3.5}\footnote{https://chat.openai.com/} \citep{DBLP:journals/corr/abs-2304-11633,DBLP:journals/corr/abs-2305-14450} is a groundbreaking conversational LLM developed by OpenAI. Researchers assess the information extraction capabilities of ChatGPT3.5 from many perspectives systematically. Since the code is not open-source, all reports are based on zero-shot setting.}
\item{\textbf{InstructUIE} \citep{DBLP:journals/corr/abs-2304-08085} is an end-to-end LLM framework for universal information extraction, which harnesses FlanT5-11B\footnote{https://huggingface.co/google/flan-t5-xxl} as the backbone.}
\end{itemize}

\subsection{Implementation Details}
In order to make a fair comparison with the four unified methods, we leverage T5-base\footnote{https://huggingface.co/google-t5/t5-base} \citep{2020t5} and FlanT5-3B\footnote{https://huggingface.co/google/flan-t5-xl} \citep{https://doi.org/10.48550/arxiv.2210.11416} (due to resource constraint). According to transfer learning configuration, we pre-train the model on 11 source datasets, and then fintune on the target dataset. In light of the randomness in instruction selection, we fix the seed, repeat the experiment five times and average the outcomes as the reported results. For each dataset, we train on the training set, the reported results on the test set are derived from the checkpoint that yields the best performance on the development set. Except for the results of our method, all other data is recorded from the original papers of the baselines. 
% Specific experimental settings can be found in the Appendix~\ref{appd: settings}.

\begin{table*}[t!] 
\renewcommand\arraystretch{1.2} 
\centering 
\scriptsize
% \resizebox{\textwidth}{!}{\Huge
\setlength{\tabcolsep}{0.5mm}{
\begin{tabular}{l|*{3}{c}|*{3}{c}|*{4}{c}|*{4}{c}}
 \hlineB{4}
& \textbf{ACE04} & \textbf{ACE05-Ent} & \textbf{CoNLL03}
& \textbf{ACE05-Rel} & \textbf{CoNLL04} & \textbf{SciERC} 
& \multicolumn{2}{c}{\textbf{ACE05-Evt}} & \multicolumn{2}{c|}{\textbf{CASIE}}  
& \textbf{14-res} & \textbf{14-lap} & \textbf{15-res} &  \textbf{16-res} \\
& \multicolumn{3}{c|}{Ent.F1}
& \multicolumn{3}{c|}{Rel.F1} 
&  Trig.F1 & Arg.F1 &  Trig.F1 & Arg.F1
& \multicolumn{4}{c}{Senti Trip.F1} \\
\hline
\texttt{TIE} (T5-base) & \textbf{87.59} & \textbf{86.42} & \textbf{92.92} & \textbf{64.44} & \textbf{73.58} & \textbf{34.41} & \textbf{73.09} & \textbf{56.71} & \textbf{74.43} & \textbf{63.14} & \textbf{75.69} & \textbf{60.36} & \textbf{66.78} & \textbf{75.17} \\
\hline
- Instruction & 86.18 & 84.22 & 91.67 & 62.87 & 71.71 & 32.79 & 71.97 & 52.19 & 68.78 & 57.77 & 73.08 & 58.55 & 64.32 & 72.93 \\
- Transfer     & 87.44 & 85.11 & 92.15 & 63.69 & 73.49 & 34.11 & 72.85 & 55.11 & 73.57 & 62.38 & 73.95 & 59.69 & 66.61 & 73.79 \\
- Regularization & 86.14 & 85.95 & 91.86 & 64.09 & 73.46 & 33.87 & 72.64 & 55.32 & 74.12 & 62.53 & 72.78 & 58.68 & 65.73 & 73.06 \\
 \hlineB{4}
    \end{tabular}}
\caption{Ablation studies for 12 IE datasets with T5-base backbone. }
\label{tab:ablation}
\end{table*}

\begin{table*}[t!] 
\centering 
\scriptsize
{
        \small
    \begin{tabular}{p{1.7cm}<{\centering} l|*{3}{c}|*{3}{c}|*{1}{c}|*{1}{c}}
       \hlineB{4}
        &  &\multicolumn{3}{c|}{\textbf{Few-Shot}} & \multicolumn{3}{c|}{\textbf{Low-Resource}} & \textbf{Full} & \multirow{2}{*}{\textbf{AVG}}\\
       \parbox[t]{1.7cm}{\textbf{Dataset}} & \textbf{Method} & \textbf{1-shot} & \textbf{5-shot} & \textbf{10-shot} & \textbf{1\%} & \textbf{5\%} &\textbf{10\%} &\textbf{100\%} \\
         \hline
       \multirow{3}{*}{\parbox[t]{1.7cm}{CoNLL03}} & UIE & 46.43 & 67.09 & 73.90 & 82.84 & 88.34 & 89.63 & 91.94 &77.16\\
                                & \texttt{TIE} (T5-base) & \textbf{49.46} & \textbf{70.62} & \textbf{74.85} & \textbf{87.22} & \textbf{89.84} & \textbf{90.16} & \textbf{92.92} & \textbf{79.29}\\
                                &w/o Transfer & 46.24 & 68.09 & 74.41 & 85.56 & 88.76 & 90.10 & 92.15 &77.90 \\
         \hline
        \multirow{3}{*}{\parbox[t]{1.7cm}{CoNLL04}}& UIE & 22.05 & \textbf{45.41} & 52.39 & 30.77 & \textbf{51.72} & 59.18 & 73.48 & \textbf{47.85}\\
                                & \texttt{TIE} (T5-base)& \textbf{22.09} &  38.08 & \textbf{52.43} & \textbf{31.32} & 49.21 & \textbf{59.28} & \textbf{73.58} & 46.57\\
                                &w/o Transfer  & 19.02 & 35.14 & 52.08 & 31.12 & 47.32 & 59.20 & 73.49 & 45.34\\
       \hline
        \multirow{3}{*}{\parbox[t]{1.7cm}{ACE05-Evt (trigger)}}& UIE & 38.14 & 51.21 & 53.23 & 41.53 & 55.70 & 60.29 & 71.33 & 53.06\\
                                & \texttt{TIE} (T5-base)& \textbf{39.12} &  \textbf{52.88} & \textbf{55.56} & \textbf{42.86} & \textbf{57.84} & 61.20 & \textbf{73.09} &\textbf{54.65}\\
                                &w/o Transfer  & 38.88 & 52.14 & 54.94 & 41.80 & 56.11 & \textbf{61.28} & 72.85 & 54.00\\

         \hline
        \multirow{3}{*}{\parbox[t]{1.7cm}{ACE05-Evt (argument)}}& UIE & 11.88 & 27.44 & 33.64 & 12.80 & 30.43 & 36.28 & 50.62 &29.01\\
                                & \texttt{TIE} (T5-base)& \textbf{12.31} &  \textbf{30.63} & \textbf{36.36} & \textbf{15.75} & \textbf{34.73} & \textbf{39.42} & \textbf{56.71} & \textbf{32.27}\\
                                &w/o Transfer  & 11.92 & 28.56 & 35.17 & 14.58 & 34.38 & 37.68 & 55.11 & 31.05\\

        \hline
        \multirow{3}{*}{\parbox[t]{1.7cm}{16-res}}& UIE & \textbf{10.50} & 26.24 & 39.11 & 24.24 & 49.31 & 57.61 & 72.86 & 39.98\\
                                & \texttt{TIE} (T5-base)& 6.860 &  \textbf{27.19} & \textbf{39.67} & \textbf{24.79} & \textbf{50.26} & \textbf{58.29} & \textbf{75.17}& \textbf{40.32}\\
                                &w/o Transfer  & 5.500 & 25.36 & 39.27 & 24.06 & 48.73 & 58.20 & 73.79 & 39.27\\

        \hlineB{4}
    \end{tabular}
    }
\caption{Results of few-shot and low-resource scenarios on four datasets. }
\label{tab:low resource}
\end{table*}

\section{Experimental Analysis}
\subsection{Main Results}
To evaluate the effectiveness of \texttt{TIE}, we compare our method with several strong baselines (Table~\ref{tab:main}).
From the results, we can get the following conclusions. \textbf{First}, \textit{there is an advantage of \texttt{TIE} by deploying the conventional language model.} \texttt{TIE} with backbone T5-base exceeds on 10 out of 12 datasets, comparing to six task-specific methods and three unified IE methods (except InstructUIE, which bases on LLM). In contrast to UIE (T5-base), there is an average improvement of 2.09 points. This result demonstrates the effectiveness of our approach. \textbf{Second}, \textit{in LLM setting, \texttt{TIE} still holds an edge.} When changing our backbone with LLM FlanT5-3B, we also outperform InstructUIE (FlanT5-11B) on three out of four common datasets. Besides, \texttt{TIE} (FlanT5-3B) achieves new state-of-the-art on almost all datasets. \textbf{Third}, \textit{the classification method proves to be more potent than the generative one in IE tasks.} The poor performance of ChatGPT3.5 in zero-shot setting proves that there are limitations in using a conversational generative model for information extraction tasks. Although the parameter scale of InstructUIE is approximately four times larger than ours, we still maintain a lead on most of the tasks, which indicates the great potential of using the classification models for information extraction (Appendix~\ref{app: generative vs classification}).

% Because the Relation F1 of InstructUIE does not consider the entity types, for a fair comparison,  we change the relation metric and then re-experiment on dataset CoNLL04 and SciERC. The result can be found inTable~\ref{tab:chaged relation f1}.

% \begin{table}[t!] 
% \centering 
% \scriptsize
% % \setlength{\tabcolsep}{0.5mm}{
% {
%         \small
%     \begin{tabular}{l|*{2}{c}}
%        \hlineB{4}
%          & \textbf{CoNLL04} & \textbf{SciERC}\\
%         \hline
%          InstructUIE & 78.48 & 45.15\\
%          \hline
%           Ours {\tiny ({FlanT5-3B})} & 74.62 & 46.53\\

%         \hlineB{4}
%     \end{tabular}
%     }
% \caption{Results on two relation datasets. The expreriments use the unstrict relation f1 which neglects the entity types as the metric function. }
% \label{tab:chaged relation f1}
% \end{table}

\begin{figure*}[!t]
%\tiny
\centering
\subfigure[CoNLL03] {\includegraphics[scale=0.3]{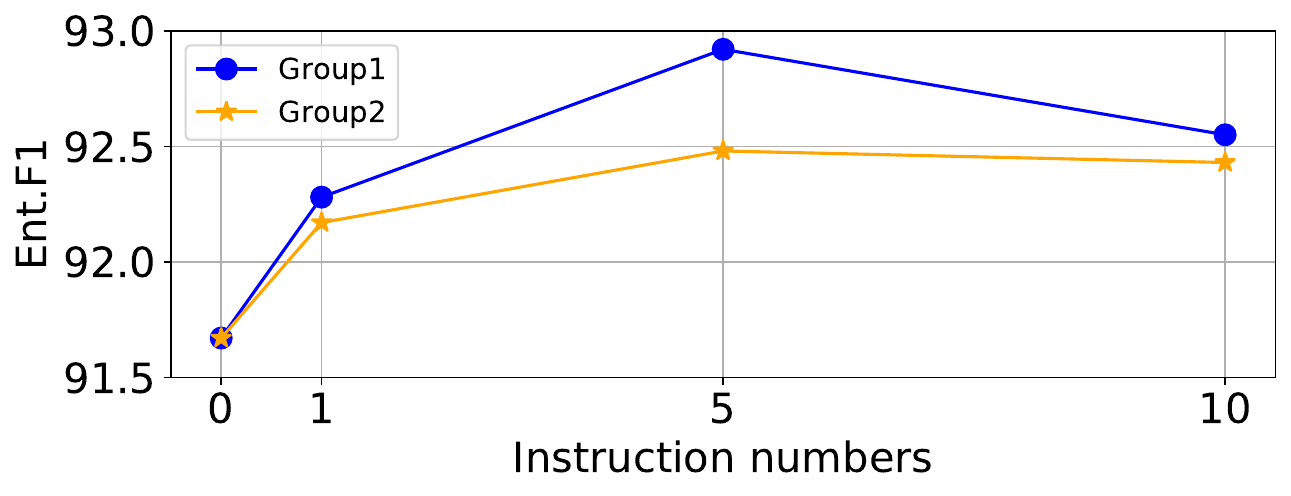}}
\subfigure[CoNLL04] {\includegraphics[scale=0.3]{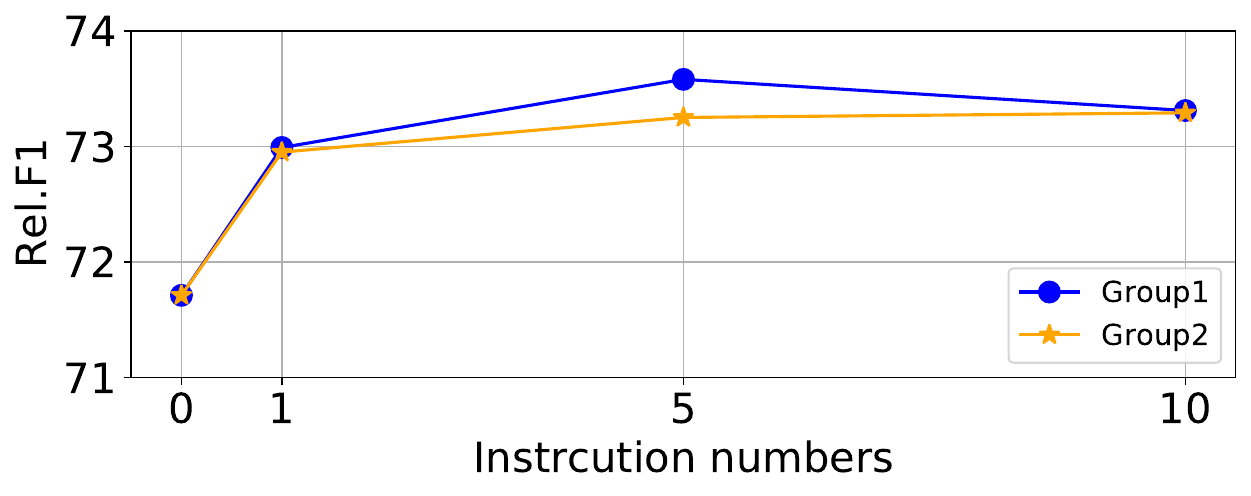}}
% \subfigure[] {\includegraphics[scale=0.235]{figures/output_figure3.pdf}}
\subfigure[ACE05-Evt (argument)] {\includegraphics[scale=0.3]{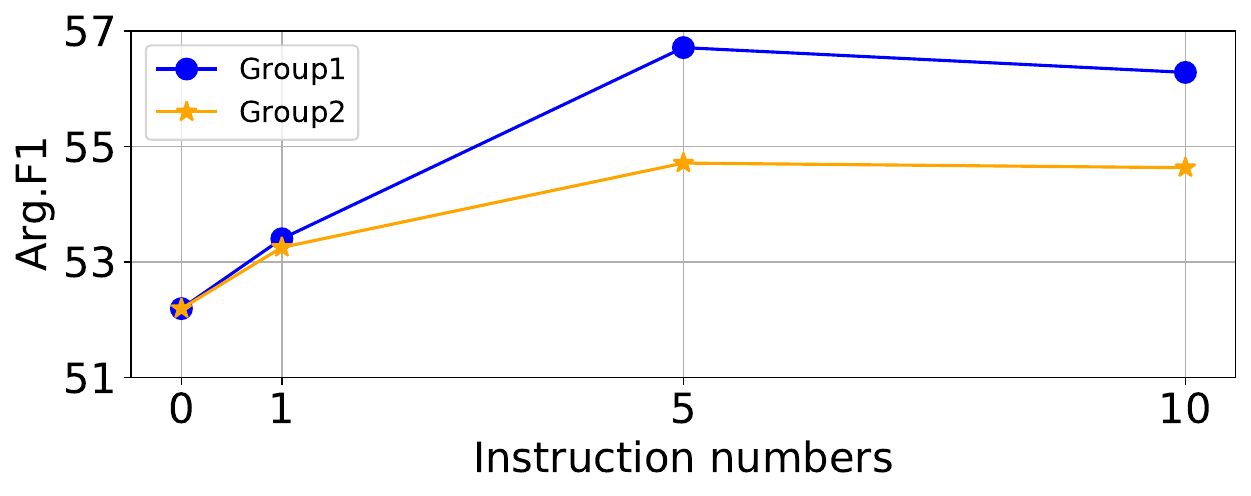}}
\subfigure[16-res] {\includegraphics[scale=0.3]{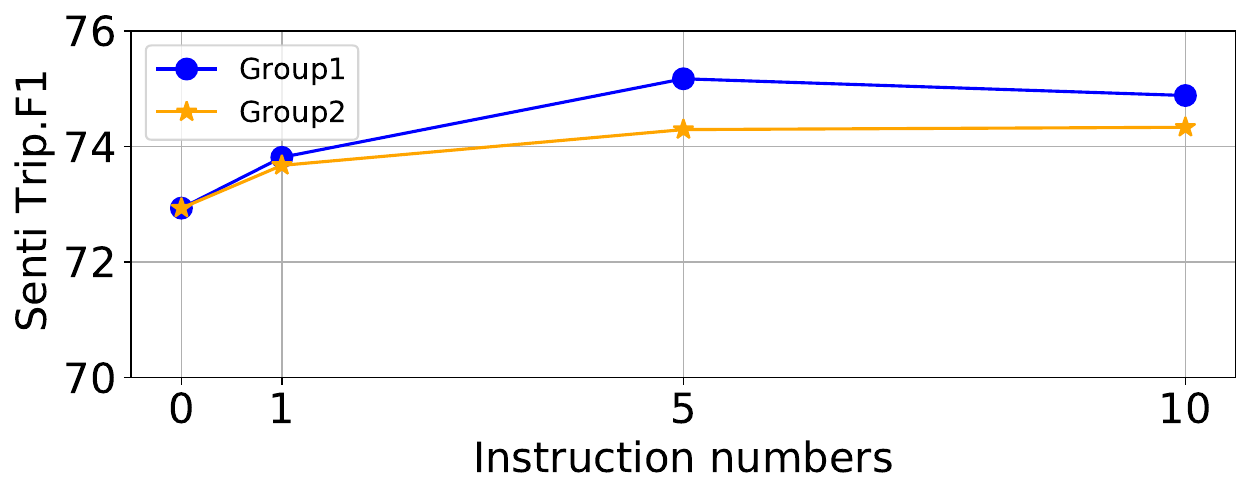}}
\caption{The influence of instruction numbers and the syntactic diversity of instructions.
}
  \label{fig:instruction}
\end{figure*}

\subsection{Ablation Studies}
For ablation studies, we experiment on each dataset of different IE tasks and study the effect of different components of our method (Table~\ref{tab:ablation}).
 `- Instruction' removes the decoder part with instructions from \texttt{TIE}.
Whereas `- Transfer' means we train our model on target datasets without transfer, which means we conduct single-task learning.
`- Regularization' stands for the absence of task-specific regularization while retaining transfer. 

We have several observations. \textbf{First}, \textit{the instructions are most important to \texttt{TIE}.} It decreases by an average of 2.55 F1 without instructions. Despite transfer learning can excavate commonality, it is the instructions that help learn a wealth of common knowledge from various tasks. Instructions inform the model of the label information to provide proper guidance, thereby alleviating the disruption.
\textbf{Second}, \textit{the gradient-regularization strategy plays a role in resolving inconsistency.} Removing task-specific regularization leads to an average decrease of 1.036 points. However, there is only a decrease of 0.771 F1 averagely when we conduct single-task learning.
For the inconsistency and complexity of knowledge, direct transfer learning does not significantly help model perform IE tasks. In order to achieve a balance between tasks, the regularization is indispensable.

\subsection{Results on Data-scarce Scenarios}
As shown in Table~\ref{tab:low resource}, we conduct experiments on 4 datasets of different IE subtasks in data-scarce scenarios. \texttt{TIE} averagely improves the F1 for 2.13, 1.59, 3.26, 0.34, compared to UIE \citep{DBLP:conf/acl/0001LDXLHSW22} on CoNLL03, ACE05-Evt (Trig.F1), ACE05-Evt (Arg.F1) and 16-res, respectively. Similarly, we remove the transfer learning step. We observe that: \textit{It is the transfer learning by gradient-regularization that fosters the effectiveness of the model in data-scarce scenarios.} The effectiveness in data-scarce scenarios demonstrates that the inconsistent knowledge resolved by regularization is negative in IE tasks. Without a large volume of training corpus, \texttt{TIE} can still acquire rich semantic information from labels within instructions. These results reveal that \texttt{TIE} has good generalization performance and is highly sensitive to new data. 

% \begin{figure*}[!t] 
% \centering
%     \includegraphics[width=0.45\textwidth]{figures/output_figure1.pdf} 
%     \includegraphics[width=0.45\textwidth]{figures/output_figure2.pdf} 
%     % \includegraphics[width=0.3\textwidth]{figures/output_figure3.pdf} 
%     \includegraphics[width=0.45\textwidth]{figures/output_figure4.pdf} 
%     \includegraphics[width=0.45\textwidth]{figures/output_figure5.pdf}
%     \caption{} 
%     \label{fig:instruction number}
% \end{figure*}

\subsection{Analyses on Instruction Diversity}
To investigate the impact of instructions on model learning, we conduct experiments on four datasets: CoNLL03, CoNLL04, ACE05-Evt (Arg.F1) and 16-res with the following setups: 
\begin{itemize}[leftmargin=*, align=left]
    \item First, we investigate the influence of the number of instructions;
    \item Second, we explore the influence of syntax diversity by customizing instructions with diverse syntactic similarity for each dataset. We generate instructions with varying syntactic similarity using ChatGPT3.5 and score them with GPT4.
  \end{itemize}

\paragraph{The Influence of Instruction Numbers}
We select instruction with quantities of 0, 1, 5 and 10 (Figure~\ref{fig:instruction}). Although \texttt{TIE} excavates commonality and resolves inconsistency via transfer learning, it still struggles when the instruction number is 0. Instructions can provide the model with elaborate task guides and label semantics, so that the decision-making would be more accurate. \texttt{TIE}'s performances on four datasets all reach the optimum when the quantity is set to 5. However, the accuracy of our method experiences a decline when employing instructions with a quantity of 10 on each dataset. More instructions could bring noise to the model, thereby decreasing the property.

\paragraph{The Influence of Syntax Diversity of Instructions}
While building the instruction pool, the prompt `... the rewritten sentences exhibit significant differences in syntax...' is input into ChatGPT3.5, for the purpose that we want to get instructions with high syntactic richness. These rephrased instructions are assigned to \textbf{Group1}. As a comparison, we perform partial word replacements in the manual instructions for each dataset, ensuring syntactical consistency. These instructions are allocated to \textbf{Group2}. 

We utilize the superior LLM GPT4 to score the syntactic richness of these instructions. The score of two groups of instructions on four datasets is illustrated in Table~\ref{tab:senmantic richness}. Instructions in Group1 have higher scores than the other group, which means they have higher syntactic richness. So that we can conclude from Figure~\ref{fig:instruction}: when the instruction number is 5, more diverse instructions lead to better model performance, while the quantity is 10, instructions possessing a greater syntactic richness could interfere with the model's learning. Nevertheless, similar instructions, whether the quantity is 5 or 10, have a limited impact on model performance. For information on the use of GPT4, please refer to the Appendix~\ref{app: gpt4}. In this way, more diverse instructions with rich syntax will be used for a specific dataset, which yields better results.

In a nutshell, we choose 5 instructions with higher syntactic richness for each dataset.

\begin{table}[t] 
\centering 
\small
\setlength{\tabcolsep}{3.0mm}{
        \small
    \begin{tabular}{l|*{2}{c}}
       \hlineB{4}
        \textbf{Dataset} & \textbf{Group1} & \textbf{Group2} \\
         \hline
       CoNLL03 & 0.88 & 0.74 \\
       CoNLL04 & 0.81 & 0.62\\
       ACE05-Evt & 0.81 & 0.72\\
       16-res & 0.86 & 0.74\\
        \hlineB{4}
    \end{tabular}
    }
\caption{Syntactic richness scored by GPT4 of two instruction groups on 4 IE datasets.}
\label{tab:senmantic richness}
\end{table}

\section{Conclusion}
In this paper, we propose a regularization-based transfer learning method for IE named \texttt{TIE}, applying an instructed graph decoder. 
It captures the shared common knowledge among tasks while preventing inconsistencies using the instructed graph decoder and the task-specific regularization strategy. Experimental results demonstrate that \texttt{TIE} achieves new state-of-the-art performance on most IE datasets, compared to both task-specific and unified baselines. 
The ablation studies show the great advantages of the main components contained in \texttt{TIE}. 
Also, we observe that \texttt{TIE} performs well on data-scarce scenarios.
In the future, it would be interesting to explore the effectiveness of our method with large-scale language models such as LLaMa and Vicuna.

\section*{Limitations}
In this paper, we propose a novel regularization-based transfer learning method for IE (named \texttt{TIE}), whose main component is a specially designed instructed graph decoder. Our method pre-trains on the source datasets and finetunes on the target one. We conduct extensive experiments on 12 datasets spanning four IE tasks, and the results demonstrate the great advantages of our proposed method in both fully supervised and data-scarce scenarios. However, there are still some limitations of our method.

(1) Although the model's structure is quite simple, the entire process of pre-training on the source datasets and finetuning on the target datasets is relatively complex and time-consuming.

(2) Due to resource limitations, we are unable to train the FlanT5-11B model, just as InstructUIE does.

(3) We do not investigate how to construct instructions that cover an open set of options. This is a very valuable area for our future work. And, for each instruction, we need to extract the label slots in the instruction sentences, which also increases the workload.

(4) For fair comparisons, our baseline data is sourced directly from their original papers. In the future, we can test our method on a wider range of models such as LLaMa and Vicuna.

\section*{Ethics Statement}
The model architecture in this paper, such as the encoder-decoder and biaffine parts, are commonly used deep learning components. All datasets mentioned in this paper are widely used public datasets in information extraction tasks. We annotate the sources of these datasets in Section 4.1, so there are no copyright issues.

All authors are knowledgeable about the research presented in this paper.

The entire process and outcomes are free from intellectual property and ethical legal disputes.

All intellectual property rights for the content of this paper belong to the authors.

\section*{Acknowledgement}
This research is funded by the National Key Research and Development Program of China (No.2021ZD0114002), the National Natural Science Foundation of China (No.62307028), the Science and Technology Commission of Shanghai Municipality Grant (No.22511105901 and No.21511100402), and Shanghai Science and Technology Innovation Action Plan (No.23ZR1441800 and No.23YF1426100). 

We still want to thank Hang Yan for providing us with some processed data. The compulsory code inspection by Jiaju Lin and Zhikai Lei also plays a significant role.

\section{Bibliographical References}
\bibliographystyle{lrec-coling2024-natbib}
\bibliography{lrec-coling2024-example}

\section{Appendices}

\subsection{The Construction of Instruction Pool}
\label{app: instruction pool}
As described in Section 3.2 (Instruction Pool), we manually write an instruction for each dataset as the seed, such as "Annotate the polarity (positive, negative or neutral), expression, aspect of the sentence." for 16-res dataset. 

Then we use the prompt "Please rewrite the following sentence several times and make sure the rewritten sentences exhibit significant differences in syntax, compared to the original sentence: Annotate the polarity (positive, negative or neutral), expression, aspect of the sentence." to augment the manual instruction. 

 Regarding to the ChatGPT3.5 settings, we utilize ChatGPT (gpt-3.5-turbo)\footnote{https://openai.com/blog/gpt-3-5-turbo-fine-tuning-and-api-updates} with four different temperatures ranging from 0.1 to 0.4. For each temperature, we generate several instructions and manually select one instruction with significant syntactic differences from the results. This process is repeated for each temperature, resulting in a total number of five instructions (including the seed) for each dataset. 

 \subsection{More Examples of the Instructions}
\label{app: examples of the instrutions}
In this section, we will display all instructions for the 14-res, 14-lap, 15-res and 16-res datasets (the ABSA task) below. All other instructions can be seen in our github repository.

1. Annotate the polarity (positive, negative or neutral), expression, aspect of the sentence.

2. Get the polarity (whether positive or negative or neutral) of the sentence, and the distinct expression and aspect of this statement.

3. Retrieve the emotional tone (positive, negative, or neutral) of the sentence and identify the corresponding content as either expression text, aspect text.

4. Find the sentiment (positive, negative, or neutral) of the sentence and identify the expression, aspect element.

5. Determine whether the sentiment of this sentence is positive, negative, or neutral and pinpoint the specific expression and aspect.

 \subsection{Explanations about the Comparison between Generation and Classification}
\label{app: generative vs classification}
 ChatGPT3.5 and InstructUIE are generative models, which generate a sequence as the extraction result directly. In contrast, our method is a classification one, which predicts the probability of each label for each position as formulated in equation (4)-(7), and utilizes the binary cross-entropy (BCE) loss to optimize the model in equation (9). 

This claim suggests that despite having significantly fewer parameters compared to these baselines as ChatGPT3.5 and InstructUIE, our method still achieves good performance on most datasets, which indicates the great potential of using the classification models for information extraction.

 \subsection{The Details on the Use of GPT4}
\label{app: gpt4}
Inspired by the effectiveness of using GPT4\footnote{https://openai.com/gpt-4} in NLG evaluation in previous studies \citep{wang-etal-2023-chatgpt,li2024leveraging}, we also utilize it to better quantify the syntax diversity of instructions. Specifically, we first use the Spacy\footnote{https://spacy.io/} library to obtain the syntactic parse trees for the instructions, and then integrate the parsing results into the prompt for scoring. 
- We will provide more details of using GPT4 for scoring the syntax diversity, including the prompt settings and the previously founded studies. We use the following prompt.
-------------------------------------------------------------------

Here are two sentences:

1.\textbf{[sentence1]}
2.\textbf{[sentence2]}

Here are the two syntactic parse trees of the sentences:

1.\textbf{[tree1]}
2.\textbf{[tree2]}

Assign a score for syntactic diversity for the two sentences on a scale of 0 to 1, where 0 is the lowest and 1 is the highest based on the Evaluation Criteria. 

Evaluation Criteria:
In the syntactic tree, each triple consists of the first element representing the word in the original text, the second element representing the headword on which the word depends, and the third element representing the dependency relationship between them.
Syntactic Diversity (0-1) - The richness of syntax between two sentences. If two sentences express the same meaning semantically but have different dependency relationships in their syntactic structures, the higher the score, the greater the difference in dependency relationships.
--------------------------------------------------------------------
 
\end{document}